\documentclass{article} 
\usepackage{spconf}

\usepackage{amsmath,amssymb,bbm}

\usepackage[breaklinks=true,letterpaper=true,bookmarks=false,hidelinks]{hyperref}
\usepackage{cleveref}
\usepackage{url}

\usepackage{graphicx}
\usepackage[caption=false]{subfig}

\usepackage{caption}

\usepackage{algorithm}
\usepackage{algorithmic}
\usepackage{color,soul}
\usepackage{booktabs}
\usepackage{array}
\newcolumntype{L}[1]{>{\raggedright\let\newline\\\arraybackslash\hspace{0pt}}m{#1}}
\newcolumntype{C}[1]{>{\centering\let\newline\\\arraybackslash\hspace{0pt}}m{#1}}
\newcolumntype{R}[1]{>{\raggedleft\let\newline\\\arraybackslash\hspace{0pt}}m{#1}}

\def\eg{\emph{e.g.~}}

\def\L{{\cal L}}

\title{{Sap}Augment: Learning A Sample Adaptive Policy for Data Augmentation}

\name{
\begin{tabular}{c}Ting-Yao Hu$^{\star}$ \qquad Ashish Shrivastava$^{\dagger}$ \qquad Jen-Hao Rick Chang$^{\dagger}$ \qquad Hema Koppula$^{\dagger}$, \\
Stefan Braun$^{\dagger}$ \qquad Kyuyeon Hwang$^{\dagger}$ \qquad Ozlem Kalinli$^{\dagger}$ \qquad Oncel Tuzel$^{\dagger}$\end{tabular}
\thanks{$^\star$Work done during summer internship at Apple. Emails: tingyaoh@andrew.cmu.edu, \{ashish.s, jenhao\_chang, hkoppula, stefan\_braun, kyuyeon, okalinli, otuzel\}@apple.com}
\vspace{-0.15in}
}
\address{
$^\star$Carnegie Mellon University \;\;\;\;\;\;\;\; $^\dagger$Apple
}
\graphicspath{{./Figures/}}

\begin{document}

\maketitle

\begin{abstract}
Data augmentation methods usually apply the same augmentation (or a mix of them) to all the training samples.
For example, to perturb data with noise, the noise is sampled from a Normal distribution with a fixed standard deviation, for all samples.
%
We hypothesize that a hard sample with high training loss already provides strong training signal to update the model parameters and should be perturbed with mild or no augmentation.
Perturbing a hard sample with a strong augmentation may also make it too hard to learn from.
Furthermore, a sample with low training loss should be perturbed by a stronger augmentation to provide more robustness to a variety of conditions.
To formalize these intuitions, we propose a novel method to learn a Sample-Adaptive Policy for Augmentation -- SapAugment.
Our policy adapts the augmentation parameters based on the training loss of the data samples.
In the example of Gaussian noise, a hard sample will be perturbed with a low variance noise and an easy sample with a high variance noise.
Furthermore, the proposed method combines multiple augmentation methods into a methodical policy learning framework and obviates hand-crafting augmentation parameters by trial-and-error.
We apply our method on an automatic speech recognition (ASR) task, and combine existing and novel augmentations using the proposed framework.
We show substantial improvement, up to $21\%$ relative reduction in word error rate on LibriSpeech dataset, over the state-of-the-art speech augmentation method.

%
\end{abstract}

\begin{keywords}
Data augmentation, speech recognition, sample-adaptive policy
\end{keywords}

\section{Introduction}
\label{sec: intro}
\begin{figure}[t]
    \centering
    \includegraphics[width=.9\linewidth]{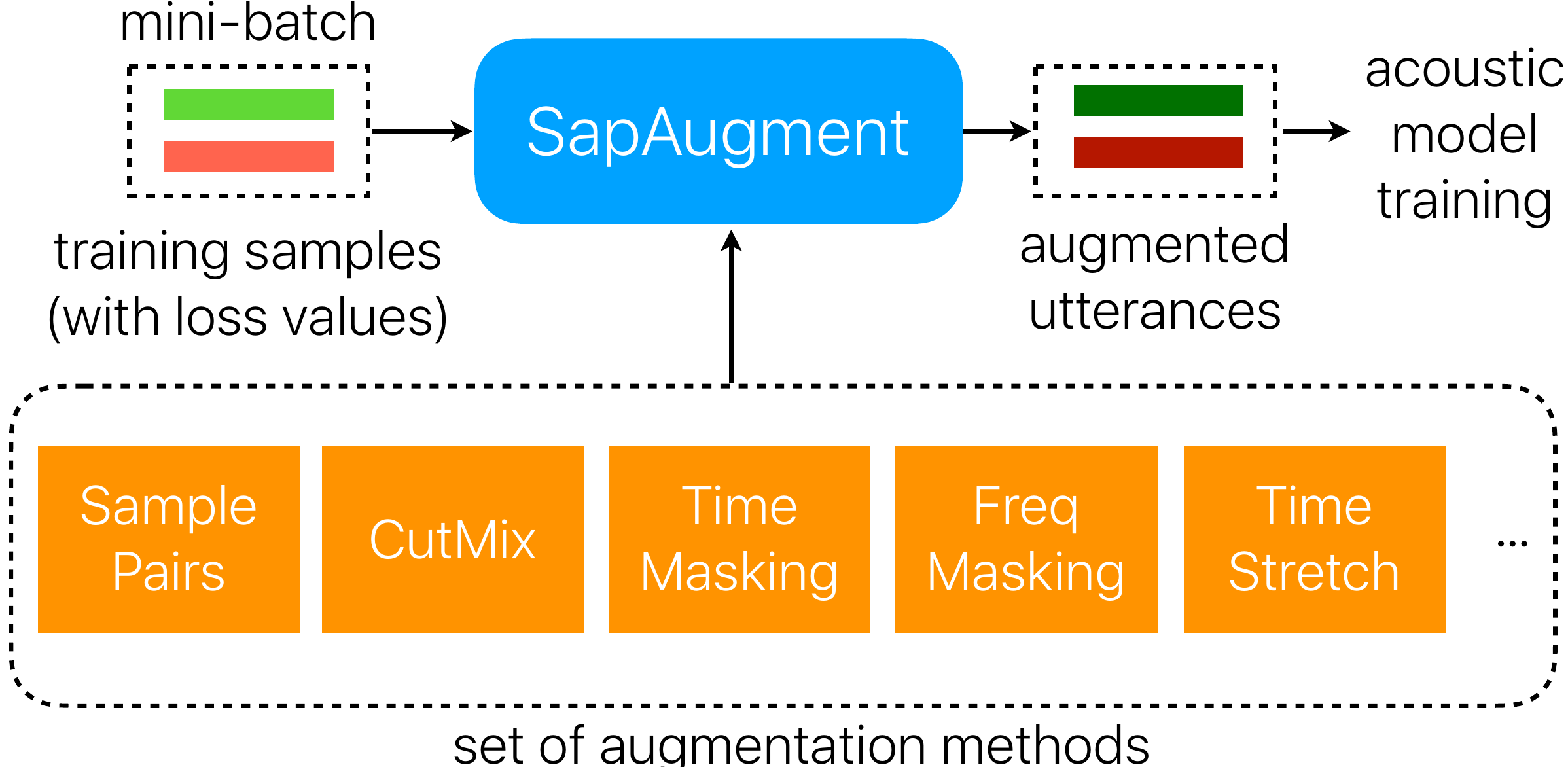} \\
    \vspace{-0.1in}
    \caption{The proposed SapAugment framework combines multiple augmentations using a learned policy that augments each training data differently using its training loss value.
}
    \label{fig:method_overview}
\end{figure}

\begin{figure}[t]
    \centering
    \includegraphics[width=.9\linewidth]{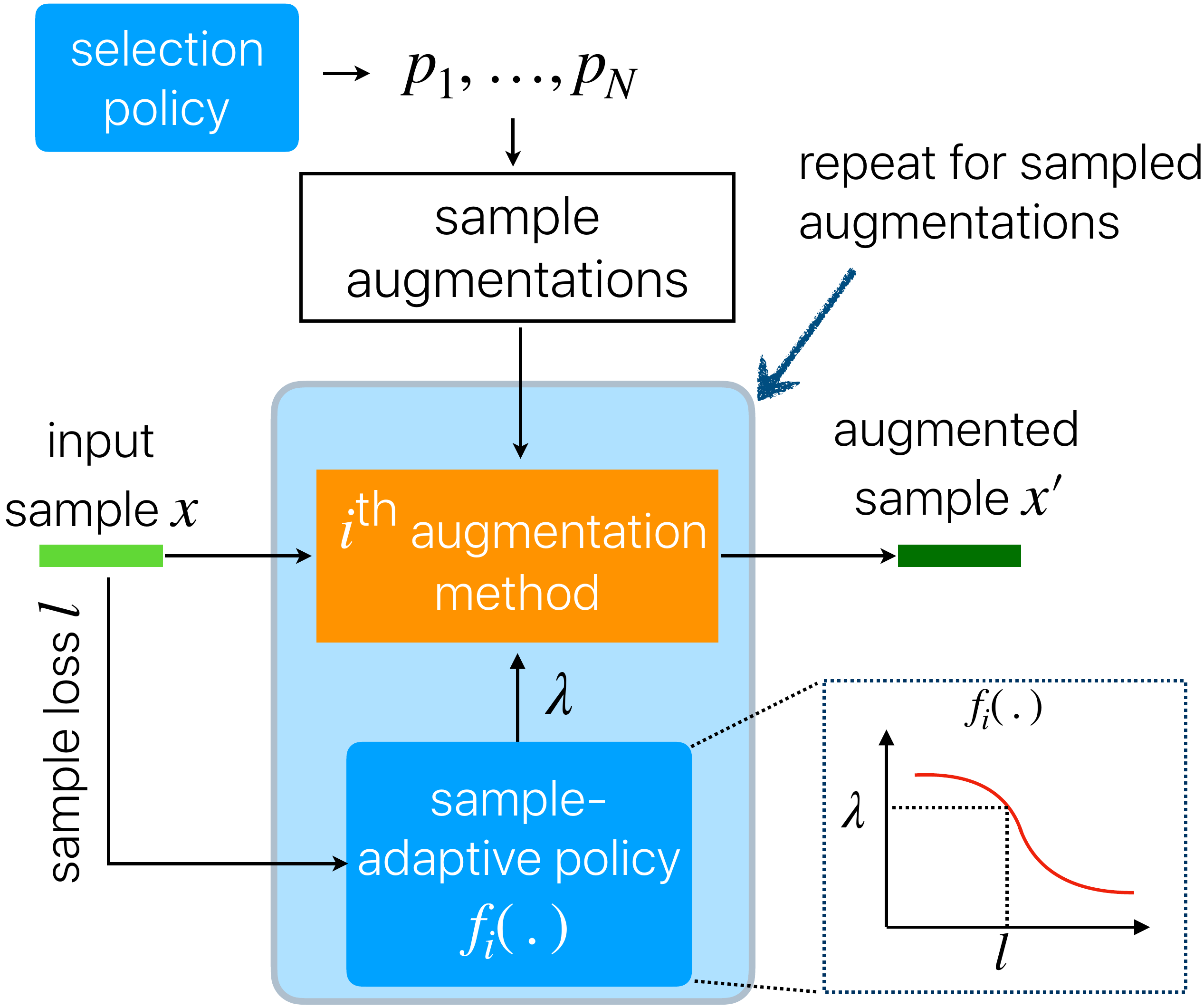} \\
    \vspace{-0.1in}
    \caption{ The selection policy provides the selection probability of the augmentation methods.
       The sample-adaptive policy, $f_i(.)$, takes training loss, $l$, as input and generates a scalar, $\lambda$, that determines the strength of the augmentation.
        A sample with low loss is augmented with a stronger augmentation, while a sample with a higher loss is augmented with a milder augmentation.  }
    \label{fig:method_detail}
\end{figure}

Data augmentation is a popular approach to reduce overfitting and improve robustness of the Deep Neural Networks (DNNs), especially when trained on a small dataset~\cite{nguyen2020improving,ko2015audio,Krizhevsky_imagenetclassification}.
While simple transformations with a hand-crafted policy are commonly used in practice, a badly-designed augmentation can harm the performance of the model~\cite{cubuk2019autoaugment, yun2019cutmix, samplepairs}.
A good augmentation strategy needs to design both the types of the data transformations (\eg masking a part of an input or adding Gaussian noise) and a policy to effectively apply these augmentations.
The policy determines the parameters of the transformation, such as the size/number of masks (for input masking), or the variance of the noise distribution (for Gaussian noise).
Furthermore, with a large variety of augmentation methods available, the policy also needs to decide the optimal subset of the augmentations for a given task.
Instead of hand-crafting the policy, one can also learn it from data~\cite{cubuk2019autoaugment,fast_autoaugment_neurips2019,randaugment_2020_CVPR_Workshops,zhang2020adversarial}, but these algorithms learn a \emph{constant} (sample-independent) policy that is applied equally to all the training samples.
However, we show that using a sample-independent policy is sub-optimal and can be improved (see Section~\ref{sec:experiment}) by adapting the policy using difficulty of the samples.
%
%
In addition, while these methods have demonstrated improvements on computer vision tasks, they have not been proven effective on speech recognition~\cite{park2019specaugment}.

The state-of-the-art augmentation method for the ASR task is a fixed augmentation policy called SpecAugment~\cite{park2019specaugment}, which perturbs the data in feature (log mel spectrogram) domain.
There are other ways of augmenting speech data, such as adding another signal of small magnitude to the input (equivalent of SamplePairing~\cite{samplepairs} for images), or replacing part of the input audio by another audio segment (similar to CutMix~\cite{yun2019cutmix} for images).
It is not obvious how to combine all of the augmentation strategies in an effective way, or find the optimal parameters of novel augmentations (such as SamplePairing and CutMix for speech).
In this paper, we propose to learn a novel \emph{sample-adaptive} policy that \emph{automatically} combines multiple augmentation methods (Figure~\ref{fig:method_overview}).
An easy sample (low training loss) is perturbed by a larger amount compared to a hard sample (high training loss).
On the ASR task, the proposed SapAugment method outperforms the current state-of-the-art data augmentation, SpecAugment~\cite{park2019specaugment}. 

%
%

%
%

%
%
We provide an overview of the method in Figure~\ref{fig:method_detail}.
We learn a sample-independent selection policy that outputs probabilities with which each augmentation is applied.
%
For each augmentation, we learn a policy parameterized by two hyper-parameters using incomplete beta function~\cite{ibeta_nist}.
Specifically, the policy takes the training loss value of a sample and outputs a scalar representing the amount of augmentation applied to the sample.
For example, for time masking, the amount of augmentation is the size of the mask applied.
The policy outputs a scalar that is larger (i.e. stronger augmentation) for a lower loss value and smaller (i.e. milder augmentation) for a larger loss value.
Since we apply the same policy to all of the mini-batches during the course of model training, it needs to account for the large variation of training losses as the model is trained.
A policy that takes the raw loss values as input would treat all the samples in the beginning of the training as hard samples (due to their high training loss from an untrained model).
Such a policy would apply mild augmentation in the beginning of the training and strong augmentation towards the end of the training.
Our goal is to learn a sample-adaptive policy that is not dependent on the state of the training.
To this end, we rank the mini-batch samples by their training losses, and instead use these loss-rankings as input to the policy.
The hyper-parameters of the policies for all augmentations and their selection probabilities
are jointly learned using Bayesian optimization~\cite{ginsbourger2008multi} by maximizing the validation accuracy, in a meta-learning framework.
Our main contributions include: 
\begin{itemize}
  \item defining a set of five augmentations for ASR -- two novel mixing augmentations complementary to the three augmentations used in the SpecAugment,
  \vspace{-0.1in}
  \item learning two policies to select and adapt parameters of the augmentation methods per training sample, and
  \vspace{-0.1in}
  \item reducing word error rate by up to $21\%$ over the state-of-the-art ASR augmentation method.
  \vspace{-0.1in}
\end{itemize}

\section{Related Works}
Speech augmentation methods perturb training data in different manners.
For example, \cite{Jaitly_vocaltract_2013} simulates vocal track perturbation, \cite{ko_audio_aug_2015} changes the speed of the audio,
\cite{Kanda2013} applies elastic spectral distortion, \cite{rir_aug_google_2017} uses room impulse responses to simulate far field data, \cite{raju2018} mixes audio at different signal to interference ratios, \cite{Ragni2014} uses a combination of augmentation schemes for a small dataset.
SpecAugment~\cite{park2019specaugment}, the state-of-the-art ASR augmentation method, applies three simple augmentations in feature (log mel spectrogram) domain -- (1) time masking for blocks of time steps, (2) frequency masking for blocks of frequency steps, and (3) time warping to randomly warp the features.
Compared to the above methods, we learn a sample-adaptive policy that combines multiple augmentation methods in a systematic manner.

Recently, a few policy based image augmentation methods have been proposed.
AutoAugment~\cite{cubuk2019autoaugment} learns a constant policy in a meta-learning framework for many image recognition tasks.
Fast AutoAugment~\cite{fast_autoaugment_neurips2019} improves the policy search time using an efficient search strategy based on density matching.
RandAugment~\cite{randaugment_2020_CVPR_Workshops} simplifies the search space of automatic policy search based augmentations without losing performance.
Adversarial AutoAugment~\cite{zhang2020adversarial} improves AutoAugment by searching a policy resulting in a higher training loss.
All of the above policy search methods learn a constant policy for all training samples.
In contrast, we learn sample-adaptive policy that perturbs the training samples based on the current loss value of the sample.
Furthermore, to the best of our knowledge, our approach is the first method to successfully apply a policy based augmentation for ASR.

\section{Sample Adaptive Data Augmentation}

\begin{figure}[t]
    \centering
    \includegraphics[width=.9\linewidth]{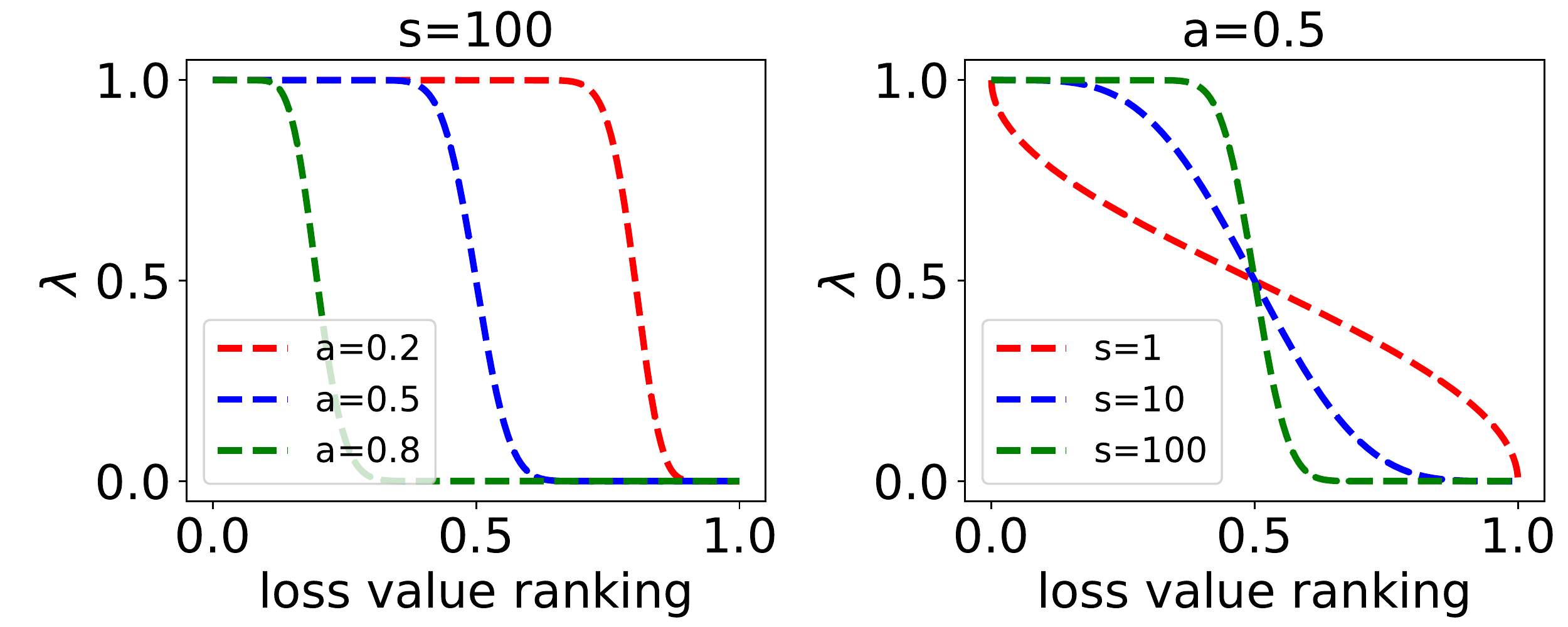} \\
    \vspace{-0.1in}
    \caption{Behavior of policy function w.r.t. the hyper-parameters $s, a$.
    The $x$-axis is the loss ranking (within a mini-batch) divided by the batch size.
    (Left) Larger $a$ results in larger $\lambda$, resulting in stronger augmentation.
    (Right) The hyper-parameter $s$ controls the slope of the mapping function.
    A larger $s$ makes the shape of $f$ closer to a step function.}
    \label{fig:ibeta_illustration}
\end{figure}
The proposed sample-adaptive policy, $f$, maps the training loss value, $l$, of a sample to a scalar, $\lambda \in [0, 1]$.
The $\lambda=f(l)$ determines the amount of augmentation applied, which, as discussed in Section~\ref{sec: intro}, needs to be larger for a smaller $l$.
We propose to use incomplete beta functions to satisfy this property.
%
The policy, $f$, is parameterized by two hyper-parameters, $s, a$, and is defined using incomplete beta function~\cite{ibeta_nist} as follows,
\begin{equation}
f_{s,a}(l) = 1 - I(s(1-a), s \cdot a; l_{rank}/B ),
\label{eq:policy_form}
\end{equation}
where $l_{rank}=1,2,...,B$ is the ranking of the loss value $l$ in a mini-batch, $B$ is the mini-batch size, and $I(\alpha, \beta; x)$ is the incomplete beta function~\cite{ibeta_nist}.
To standardize the input to the policy, we use ranking of the loss instead of the raw loss value in Equation~\eqref{eq:policy_form}.
The range of normalized loss ranking, $l_{rank}/B$, is from $0$ to $1$, while the raw training loss values change significantly during the course of training.
A few samples of the policy with different values of $s, a$ are illustrated in Figure~\ref{fig:ibeta_illustration}.
Learning a policy for an augmentation is equivalent to learning the hyper-parameters, $s, a$, for that augmentation.
Furthermore, we apply the augmentation policy, $f$, only with probability $p$ such that we do not augment the data using this policy with probability $1-p$.
This selection probability, $p$, is also learned along with the policy hyper-parameters, $s, a$.

\subsection{Augmentation methods used in SapAugment}
SapAugment combines $N$ augmentation methods by jointly learning $N$ policies, $f_{s_1,a_1}, \dots, f_{s_N, a_N}$ and their selection probabilities, $p_1, \dots, p_N$.
In this work, we combine $5$ augmentation methods, $3$ in the feature (log mel spectrogram) domain and $2$ in raw speech domain.
\subsubsection{Feature domain augmentations}
\begin{itemize}
\item \emph{Time masking} and \emph{frequency masking} were originally introduced in SpecAugment \cite{park2019specaugment}.
The two methods simply mask out segments of sizes $m_t$ and $m_f$ along the time and the frequency axes, respectively, in log mel spectrogram feature.
In our implementation, the masked region is filled with the average values along the masked dimensions.
For both masking schemes, the number of masks is kept constant ($4$ in our experiments).
\item \emph{Time stretching}~\cite{nguyen2020improving, ko2015audio} re-samples the feature sequence along the time axis by a fix ratio $\rho$, effectively changing the speaking rate in the signal.
Given speech features for a sequence of $T$ frames, $\mathbf f_0, \dots, \mathbf f_{T-1}$, time stretching generates $(1+\rho) T $ features, $\mathbf f'_0, \dots, \mathbf f'_{(1+\rho)T-1}$, such that  $\mathbf f'_i = \mathbf f_{\lfloor i/(1+\rho) \rfloor}, i=0,1,..., \lfloor (1+\rho)T \rfloor -1$.
The stretching ratio $\rho$ is sampled from a uniform distribution $U(-\rho_0, \rho_0)$


\end{itemize}

\subsubsection{Raw speech domain augmentations}
\label{subsec: raw speech augmentations}
\begin{itemize}
\item \emph{SamplePairing} (SP)~\cite{samplepairs}, originally developed for image augmentation, mixes $i^{\text{th}}$ input, $\boldsymbol x_i$, with another speech $\boldsymbol x_j$ in the training set:
$\boldsymbol x_i' = (1-\lambda_{sp}) \boldsymbol x_i + \lambda_{sp} \boldsymbol x_j,$
where $\boldsymbol x_i'$ is the augmented speech and $\lambda_{sp}$ is the combination weight.
Since the duration of $\boldsymbol x_i$ and $\boldsymbol x_j$ may differ, we pad (by repeating) or clip $\boldsymbol x_j$ as needed.

\item \emph{CutMix}~\cite{yun2019cutmix}, originally proposed for images, replaces part of an input speech $\boldsymbol x_i$ by segments from another training sample $\boldsymbol x_j$:
\begin{equation*}
\boldsymbol x_i[t_{i,k}:t_{i,k} + w] \leftarrow \boldsymbol x_j[t_{j,k}:t_{j,k} + w], k=1,2,...,N_{cm},
\end{equation*}
where $N_{cm}$ is the total number of replaced segments (set to $6$ in our experiments), and $w$ is the size of the segments.
Unlike the original CutMix for image augmentation, we do not change the label information of the perturbed sample.
\end{itemize}

Each augmentation method contains one parameter to control the augmentation magnitude, as listed in Table~\ref{tab:magnitude_control}.
The value of $\lambda \in [0, 1]$ from the policy is mapped to this augmentation parameter by a simple $1$-d affine transformation, as shown in the last column of the Table.

\begin{table}[t]
    \centering
\begin{tabular}{@{}C{2.9cm}|C{1.9cm}|C{2.78cm}@{}}
  \toprule[0.015in]
          aug. parameter      & range & mapping from $\lambda$                           \\ \midrule
       $m_t$ (time masking)                                                                       & \{2,3,4,5,6\}              & $m_t^*= \lfloor 2+4\lambda \rfloor$        \\
 $m_f$ (freq. masking )                                                             & \{2,3,4,5,6\}              & $m_f^*= \lfloor 2+4\lambda \rfloor $        \\
    $\rho_0$ (time stretching)                                                                       & {[}0.2,0.6{]}              & $\rho_0^*= 0.2+0.4\lambda$ \\
      $\lambda_{sp}$   (SamplePairing)                                                                 & {[}0,0.1{]}                & $\lambda_{sp}^*=0.1\lambda$               \\
      $w$  (CutMix)                                                                               & {[}1600, 4800{]}           & $w^*=1600+3200\lambda$         \\ \bottomrule
\end{tabular}
\caption{The augmentation parameter values, chosen from the above range, determine the amount of augmentation.
For a given input sample, the learned policy chooses one of the values in the range above.
For CutMix, the range of $w$ correspond to $[0.1, 0.3]$ seconds at $16$kHz sampling rate of audio.
}
\label{tab:magnitude_control}
\end{table}

\subsection{Policy search}
The training of SapAugment framework is formulated as finding a policy to select and adapt augmentation parameters that maximizes the validation accuracy.
The policy search space is the set of policy hyper-parameters for all the augmentations, $\boldsymbol v_p = [s_1,a_1,s_2,a_2,...,s_N, a_N]$, and their selection probabilities, $\boldsymbol p = [p_1, \dots, p_N]$, where $N$ is the number of augmentation methods.
We use Bayesian optimization \cite{ginsbourger2008multi}, which uses a Gaussian process to model the relation between $(\boldsymbol v_p, \boldsymbol p)$ and the validation accuracy.
To further accelerate the search process, we utilize constant liar strategy to parallelize the suggestion step of Bayesian optimization.
Please refer to the Algorithm~$2$ of \cite{ginsbourger2008multi} for more details.


%
\section{Experiments}
\label{sec:experiment}


\begin{table}[t]
  \centering
\begin{tabular}{@{}C{1.6cm}C{1.9cm}C{1.9cm}C{1.9cm}@{}}
  \toprule[0.015in]
  Dataset & No augmentation & SpecAugment~\cite{park2019specaugment} & SapAugment \\ \midrule
   \multicolumn{4}{l}{\hspace{-0.1in}LibriSpeech 100h}\\\hline
   ~~test-clean & $12.0\pm0.30$  & $10.6\pm0.20$ & {$\mathbf {8.3\pm0.05}$} \\
   ~~test-other &$30.9\pm0.35$  & $24.5\pm0.10$  & $ \mathbf {21.5\pm0.10}$ \\ \midrule
 \multicolumn{4}{l}{\hspace{-0.1in}LibriSpeech 960h}\\\hline
    ~~test-clean & $4.4\pm0.05$ & $3.9\pm0.10$  & $\mathbf {3.5\pm0.05}$\\
    ~~test-other & $11.5\pm0.10$  & $9.4\pm0.10$     & $\mathbf {8.5\pm0.10}$\\
 %
 \bottomrule
\end{tabular}
\caption{Comparison of SapAugment against baselines on the LibriSpeech dataset. All numbers are percentage word error rate (WER) (lower is better). }
\label{tab:sota}
\end{table}

\begin{table}[t]
\begin{tabular}{@{}L{5.7cm}|C{0.97cm}C{0.97cm}@{}}
  \toprule[0.015in]
 Augmentation/policy & test-clean & test-other\\  \midrule
No aug. & $12.0$ & $30.9$\\ 
CutMix aug. without policy & $11.2$ & $26.6$\\ 
SamplePairing aug. without policy & $10.2$ & $26.3$\\ 
All aug. without policy & $9.6$ & $22.3$\\
All aug. with selection policy & $9.4$             & $22.2$\\ 
All aug. with sample-adaptive policy  & $9.3$ & $21.9$\\ 
SapAugment: All aug. with sample-adaptive and selection policies  & $\mathbf{8.3}$ & $\mathbf{21.5}$\\
\bottomrule
\end{tabular}
\caption{Ablation study of SapAugment on LibriSpeech 100h training dataset.
All numbers are percentage word error rate (WER).
First row is the baseline WER without any augmentation.
Rows $2$-$4$ apply various combination of augmentations without any learned policy.
Row $5$ applies a constant policy without considering the training loss value to determine the amount of augmentation.
Row $6$ applies a sample-adaptive policy but does not include the policy selection probability, i.e. all the augmentations are always applied, only their magnitude is controlled by the policy using sample loss value.
Last row is SapAugment using all augmentations and both policies.}
\label{tab:ablation}
\end{table}

To evaluate our method, we use the LibriSpeech~\cite{panayotov2015librispeech} dataset that contains $1,000$ hours of speech from public domain audio books.
Following the standard protocol on this dataset, we train acoustic models by using both the full training dataset (LibriSpeech 960h), and the subset with clean, US accent speech only (LibriSpeech 100h).
Results are reported on the two official test sets - (1)~test-clean contains clean speech, and (2)~test-other contains noisy speech.

\subsection{Implementation}

For the network architecture of an end-to-end acoustic model, we choose the Transformer \cite{vaswani2017attention} because of its efficiency in training.
We utilize the Transformer acoustic model implemented in ESPNet \cite{watanabe2018espnet}.
Following the original implementation in \cite{watanabe2018espnet}, we use the ``small" configuration of Transformer  for LibriSpeech 100h, and the ``large" configuration for LibriSpeech 960h.
All of the model checkpoints are saved during the training process, and the final model is generated by averaging the $5$ models with highest validation accuracy.
For the acoustic feature, we apply $80$-dimensional filter bank features.

\subsection{Evaluation}
In Table~\ref{tab:sota}, we compare the proposed SapAugment to the baseline model without any augmentation, and the model trained with SpecAugment~\cite{park2019specaugment}, the current state-of-the-art augmentation method in ASR.
For a fair comparison, we ran all experiments using the same ESPNet model on all baselines and report results without the language model to isolate the improvements from data augmentation.
The results show that the proposed SapAugment outperforms SpecAugment -- up to $21\%$ relative WER reduction with LibriSpeech 100h and $10\%$ relative WER reduction with LibriSpeech 960h.
The improvement in WER is more significant in the smaller scale dataset (LibriSpeech 100h) because augmentations reduce overfitting which is more severe for a small dataset.

\subsection{Ablation Study}
To study the contribution of each component of SapAugment, we present an ablation study on the LibriSpeech 100h dataset in Table~\ref{tab:ablation}.
From this study, we make the following observations:
First, the proposed two augmentations in the raw speech domain help improve the performance compared to no augmentation.
Second, learning a policy gives better results than no policy.
Third, learning both the selection policy and the sample-adaptive policy is better than either one of them.

%
%

\section{Conclusion}

We proposed a novel data-augmentation framework that automatically adapts the strength of the perturbation based on the loss values of the samples.
The method achieved state-of-the-art data augmentation performance on ASR.
Although the method is demonstrated on speech domain, the policy learning framework is generic and could be applied to other modalities.
%
%


\begin{thebibliography}{10}

\bibitem{nguyen2020improving}
Thai-Son Nguyen, Sebastian St{\"u}ker, Jan Niehues, and Alex Waibel,
\newblock ``Improving sequence-to-sequence speech recognition training with
  on-the-fly data augmentation,''
\newblock in {\em Proc. ICASSP}. IEEE, 2020, pp. 7689--7693.

\bibitem{ko2015audio}
Tom Ko, Vijayaditya Peddinti, Daniel Povey, and Sanjeev Khudanpur,
\newblock ``Audio augmentation for speech recognition,''
\newblock in {\em Sixteenth Annual Conference of the International Speech
  Communication Association}, 2015.

\bibitem{Krizhevsky_imagenetclassification}
Alex Krizhevsky, Ilya Sutskever, and Geoffrey~E. Hinton,
\newblock ``Imagenet classification with deep convolutional neural networks,''
\newblock in {\em Proc. NIPS}, p. 2012.

\bibitem{cubuk2019autoaugment}
Ekin~D Cubuk, Barret Zoph, Dandelion Mane, Vijay Vasudevan, and Quoc~V Le,
\newblock ``Autoaugment: Learning augmentation strategies from data,''
\newblock in {\em Proc. CVPR}, 2019.

\bibitem{yun2019cutmix}
Sangdoo Yun, Dongyoon Han, Seong~Joon Oh, Sanghyuk Chun, Junsuk Choe, and
  Youngjoon Yoo,
\newblock ``Cutmix: Regularization strategy to train strong classifiers with
  localizable features,''
\newblock in {\em Proc. ICCV}, 2019, pp. 6023--6032.

\bibitem{samplepairs}
Hiroshi Inoue,
\newblock ``Data augmentation by pairing samples for images classification,''
\newblock {\em arXiv preprint arXiv:1801.02929}, 2018.

\bibitem{fast_autoaugment_neurips2019}
Sungbin Lim, Ildoo Kim, Taesup Kim, Chiheon Kim, and Sungwoong Kim,
\newblock ``Fast autoaugment,''
\newblock in {\em Proc. NeurIPS}. 2019.

\bibitem{randaugment_2020_CVPR_Workshops}
Ekin~D. Cubuk, Barret Zoph, Jonathon Shlens, and Quoc~V. Le,
\newblock ``Randaugment: Practical automated data augmentation with a reduced
  search space,''
\newblock in {\em Proc. CVPR Workshops}, 2020.

\bibitem{zhang2020adversarial}
Xinyu Zhang, Qiang Wang, Jian Zhang, and Zhao Zhong,
\newblock ``Adversarial autoaugment,''
\newblock in {\em International Conference on Learning Representations}, 2020.

\bibitem{park2019specaugment}
Daniel~S Park, William Chan, Yu~Zhang, Chung-Cheng Chiu, Barret Zoph, Ekin~D
  Cubuk, and Quoc~V Le,
\newblock ``Specaugment: A simple data augmentation method for automatic speech
  recognition,''
\newblock {\em Proc. Interspeech}, 2019.

\bibitem{ibeta_nist}
R.~B. Paris,
\newblock {\em Incomplete beta functions},
\newblock NIST Handbook of Mathematical Functions, Cambridge University Press,
  2010.

\bibitem{ginsbourger2008multi}
David Ginsbourger, Rodolphe Le~Riche, and Laurent Carraro,
\newblock ``A multi-points criterion for deterministic parallel global
  optimization based on gaussian processes,''
\newblock 2008.

\bibitem{Jaitly_vocaltract_2013}
Navdeep Jaitly and Geoffrey~E. Hinton,
\newblock ``Vocal tract length perturbation (vtlp) improves speech
  recognition,''
\newblock in {\em Proc. ICML Workshop on Deep Learning for Audio, Speech and
  Language}, 2013.

\bibitem{ko_audio_aug_2015}
Tom Ko, Vijayaditya Peddinti, Daniel Povey, and Sanjeev Khudanpur,
\newblock ``Audio augmentation for speech recognition.,''
\newblock in {\em Proc. Interspeech}, 2015.

\bibitem{Kanda2013}
Naoyuki Kanda, Ryu Takeda, and Yasunari Obuchi,
\newblock ``Elastic spectral distortion for low resource speech recognition
  with deep neural networks,''
\newblock in {\em 2013 IEEE Workshop on Automatic Speech Recognition and
  Understanding}, 2013.

\bibitem{rir_aug_google_2017}
Chanwoo Kim, Ananya Misra, Kean Chin, Thad Hughes, Arun Narayanan, Tara
  Sainath, and Michiel Bacchiani,
\newblock ``Generation of large-scale simulated utterances in virtual rooms to
  train deep-neural networks for far-field speech recognition in google home,''
\newblock in {\em Proc. Interspeech}, 2017.

\bibitem{raju2018}
Anirudh Raju, Sankaran Panchapagesan, Xing Liu~Arindam Mandal, and Nikko Strom,
\newblock ``Data augmentation for robust keyword spotting under playback
  interference,''
\newblock {\em arXiv preprint arXiv:1808.00563}, 2018.

\bibitem{Ragni2014}
Anton Ragni, Kate~M Knill, Shakti~P Rath, and Mark J.~F. Gales,
\newblock ``Data augmentation for low resource languages,''
\newblock in {\em Proc. Interspeech}, 2014.

\bibitem{panayotov2015librispeech}
Vassil Panayotov, Guoguo Chen, Daniel Povey, and Sanjeev Khudanpur,
\newblock ``Librispeech: an asr corpus based on public domain audio books,''
\newblock in {\em Proc. ICASSP}, 2015.

\bibitem{vaswani2017attention}
Ashish Vaswani, Noam Shazeer, Niki Parmar, Jakob Uszkoreit, Llion Jones,
  Aidan~N Gomez, {\L}ukasz Kaiser, and Illia Polosukhin,
\newblock ``Attention is all you need,''
\newblock in {\em Advances in neural information processing systems}, 2017, pp.
  5998--6008.

\bibitem{watanabe2018espnet}
Shinji Watanabe, Takaaki Hori, Shigeki Karita, Tomoki Hayashi, Jiro Nishitoba,
  Yuya Unno, Nelson {Enrique Yalta Soplin}, Jahn Heymann, Matthew Wiesner,
  Nanxin Chen, Adithya Renduchintala, and Tsubasa Ochiai,
\newblock ``{ESPnet}: End-to-end speech processing toolkit,''
\newblock in {\em Proc. Interspeech}, 2018.

\end{thebibliography}

\end{document}